# FGN: Fusion Glyph Network for Chinese Named Entity Recognition

Zhenyu Xuan, Rui Bao and Shengyi Jiang[*]

School of Information Science and Technology, Guangdong University of Foreign Studies, Guangdong, China
`xuanzhenyu@foxmail.com, jiangshengyi@163.com`

**Abstract.** As pictographs, Chinese characters contain latent glyph information, which is often overlooked. In this paper, we propose the FGN[1], Fusion Glyph Network for Chinese NER. Except for encoding glyph information with a novel CNN, this method may extract interactive information between character distributed representation and glyph representation by a fusion mechanism. The major innovations of FGN include: (1) a novel CNN structure called CGS-CNN is proposed to capture glyph information and interactive information between the neighboring graphs. (2) we provide a method with sliding window and attention mechanism to fuse the BERT representation and glyph representation for each character. This method may capture potential interactive knowledge between context and glyph. Experiments are conducted on four NER datasets, showing that FGN with LSTM-CRF as tagger achieves new state-of-the-art performance for Chinese NER. Further, more experiments are conducted to investigate the influences of various components and settings in FGN.

## 1 Introduction

Named entity recognition (NER) is generally treated as sequence tagging problem and solved by statistical methods or neural networks. In the field of Chinese NER, researches generally adopt character-based tagging strategy to label named entities [1, 2]. Some researches [3, 4] explicitly compared character-based methods and word-based methods for NER, confirming that character-based methods avoid the error from word segmentation stage and perform better. When using character-based methods for NER, the effect of character-level knowledge representation may greatly affect the performance of Chinese NER model.

Currently, distributed representation learning has become the mainstream method to represent Chinese characters, especially after the raise of BERT [5], which raised the baselines for almost all fields of NLP. However, these methods overlooked the infor-mation inside words or characters like Chinese glyph. There have been studies, focusing on internal components of words or characters. In English field, researchers [6] used Convolutional Neural Network (CNN) to encode the spelling of words for sequence tagging task. This method is not suitable for Chinese NER, as Chinese is not alphabetical language but hieroglyphic language. Chinese characters can be further segmented into radicals. For example, character "抓"(grasp) is constitutive of "扌"(hand) and "爪"(claw). Study on radical-based character embedding [7] confirmed the effectiveness of these components in Chinese.

---

[*] corresponding author (jiangshengyi@163.com)

[1] https://github.com/AidenHuen/FGN-NER

Further, researchers turned attention to regard Chinese characters as graphs for glyph encoding. Some researchers [8, 9, 25] tried running CNNs to capture glyph information in character graphs. However, these works just obtained neglectable improvement on trial. Avoiding the shortcomings of previous works, Meng et al. [2] proposed a glyph-based BERT model called Glyce, which achieved SOTA performances in various NLP tasks including NER. They adopted Tianzige-CNN to encode seven historical and contemporary scripts of each Chinese character. Tianzige is a traditional form of Chinese calligraphy, which conforms the radical distribution inside a Chinese character. Then Transformer [10] was used as sequence encoder in Glyce. Further, Sehanobish and Song [11] proposed a glyph-based NER model called GlyNN, which encoded only Hei Ti font of each character to offer glyph information and used BiLSTM-CRF as sequence tagger. Moreover, representations of non-Chinese characters were taken into consideration carefully in GlyNN. Compared with Glyce, GlyNN with BERT achieved comparable performance in multiple NER datasets, using less glyph resource and smaller CNN. It proved that historical scripts are meaningless for NER to some extent. We suspect this is because the types and numbers of entities in modern Chinese are far more abundant and complex than the ones in ancient times.

The above works just encoded the glyph and distributed representation independently. They ignored the interactive knowledges between glyphs and contexts, which have been studied in the field of multimodal deep learning [12, 13, 14]. Moreover, as the meaning of Chinese character is not complete, we suspect that encoding each character glyph individually is not an appropriate approach. In fact, interactive knowledge between the glyphs of neighboring characters maybe benefit the NER task. For example, characters in tree names like "杨树"(aspen), "柏树"(cypress) and "松树"(pine tree) have the same radical "木"(wood), but characters of an algorithm name "决策树"(decision tree) have no such pattern. There are more similar patterns in Chinese language, which can be differentiated by interactive knowledge between neighboring glyphs.

Therefore, we propose the FGN, Fusion Glyph Network for Chinese NER. The major innovations in FGN include: (1) a novel CNN structure called CGS-CNN, Character Graph Sequence CNN is offered for glyph encoding. CGS-CNN may capture potential information between the glyphs of neighboring characters. (2) We provide a fusion method with out-of-sync sliding window and Slice-Attention to capture interactive knowledge between glyph representation and character representation.

FGN is found to improve the performance of NER, which outperforms other SOTA models on four NER datasets (Section 4.2). In addition, we verify and discuss the influence of various proposed settings in FGN (Section 4.3).

## 2    Related Work

Our work is related to neural network for NER. Ronan et al. [15] proposed the CNN-

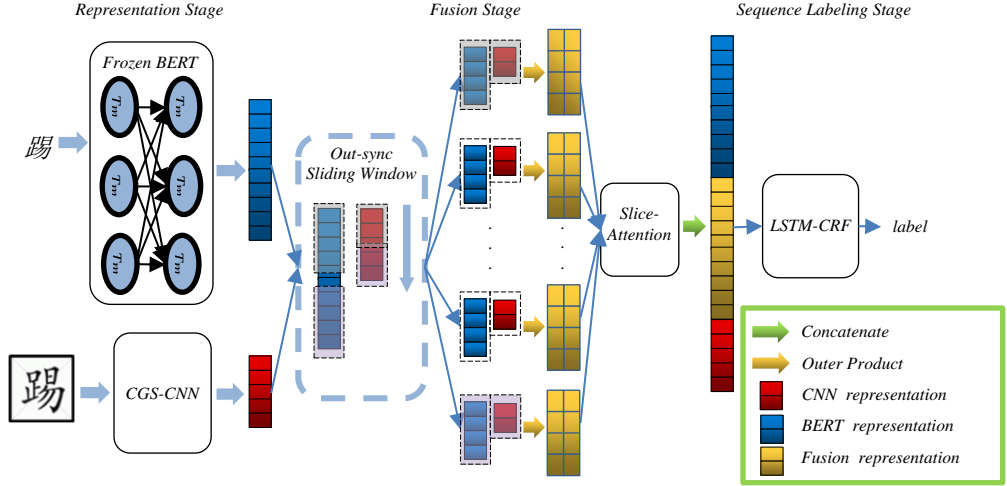

**Fig. 1.** Architecture of the FGN for named entity recognition.

CRF model, which obtained competitive performance to various best statistical NER models. LSTM-CRF [16] has been the mainstream component in subsequent NER models at present. To enhance word-level representation, Ma and Hovy [6] proposed the LSTM-CNN-CRF structure for sequence labeling, which adopted CNNs to encode the spelling of each English word for semantic enhancement. Further, a coreference aware representation learning method [17] was proposed, which was combined with LSTM-CNN-CRF for English NER. In Chinese field, Dong et al. [18] organized radicals in each character as sequence and used LSTM network to capture the radical information for Chinese NER. Zhang et al. [19] proposed a novel NER method called lattice-LSTM, which skillfully encoded Chinese characters as well as all potential words that match a lexicon. Drawing on Lattice-LSTM, Word-Character LSTM (WC-LSTM) [20] was proposed, which added word information into the start and the end characters of a word to alleviate the influence of word segmentation errors.

Our work is also related to some multimodal works. Currently, knowledge from vision has been widely-used in NLP. We simply divide these relative researches into two categories according to the source of vision knowledge: glyph representation learning and multimodal deep learning. The Former is scarce as mentioned earlier. We transform the input sentences to graph sequences for 3D encoding. To our knowledge, we are the first to encode character glyph in sentence-level by 3D convolution [21], which was mostly proposed to encode video information. The latter is current hotspot in various NLP fields. Zhang et al. [12] proposed an adaptive co-attention network for tweets NER, which adaptively balanced the fusion proportions of image representation and text representation from a tweet. With reference of BERT, a multimodal BERT [13] was proposed for target-oriented sentiment classification. Multiple self-attention layers [9] were used in this model to capture interactive information after concatenating BERT and visual representation. Further, Mai et al. [14] proposed a fusion network with local and global perspective for multimodal affective computing. They provided a sliding window to slice multimodal vectors and fused each slice pair by outer product function. And attentive Bi-directional Skip-connected LSTM was used to combine slice pairs. Our method borrows the ideas of above-mentioned methods for multimodal fusion. Different from their work that fused the sentence-level representation, we focus on character-level fusion for Chinese NER.

## 3 Model

In this section, we introduce the FGN in detail. As shown in Fig. 1, FGN can be divided into three stages: representation stage, fusion stage and tagging stage. We follow the strategy of character-based sequence tagging for Chinese NER.

### 3.1 Representation Stage

Here we discuss the representation learning for Chinese character including character representation from BERT and glyph representation form CGS-CNN. Detail of these representations are as followed.

**BERT.** BERT is a multi-layer Transformer encoder, which offers distributed representations for words or characters. We use the Chinese pre-trained BERT to encode each character in sentences. Different from the normal fine-tuning strategy, we first fine-tune BERT on training set with a CRF layer as tagger. Then freeze the BERT parameters and transfer them to FGN. experiment in Section 4.3 shows the effectiveness of this strategy.

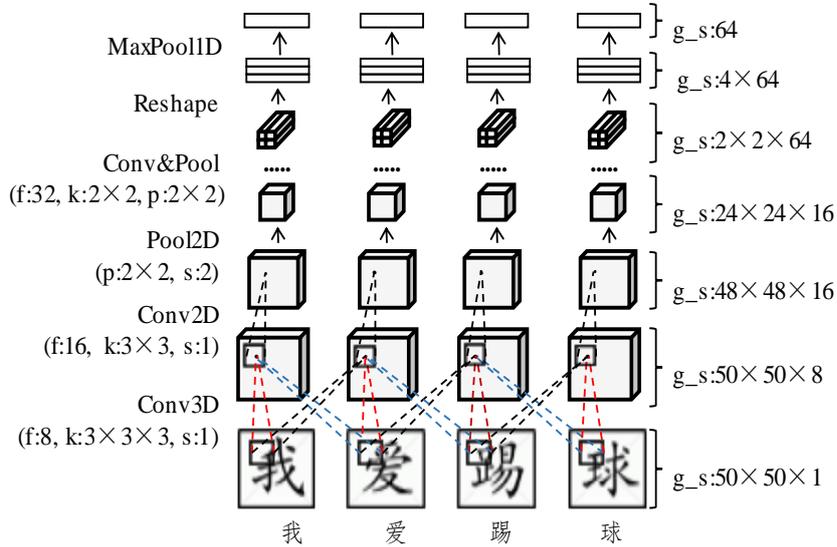

**Fig. 2.** Architecture of CGS-CNN with a input sample "我爱踢球" (I love playing football). "f", "k", "s", "p" stand for kernel number, kernel size, stride, and pooling window size. "g_s" represents the tensor size of output from each layer.

**CGS-CNN.** Fig. 2 depicts the architecture of CGS-CNN. We only choose the simple Chinese script to generate glyph vectors, as the past work [11] showed that using only one Chinese script achieved comparative performance as well as seven scripts. The input format for CGS-CNN is character graph sequence. We first convert sentences to graph sequences, in which characters are replaced with 50×50 gray-scale graphs. Characters which are not Chinese may be given corresponding initialize matrices with parameters between 0 and 1. Then we provide two 3×3×3 3D convolution layers to encode graph sequence and output each 50×50 graph with 8 channels. 3D convolution can extract feature from both spatial and temporal dimensions, which means each glyph vector may obtain additional glyph information from the neighboring graphs. Using padding on the dimension of graph sequence, we may keep the length of graph sequence constant after passing through 3D convolution, which is necessary for character-based tagging. Then the output of 3D convolution may pass through several groups of 2D convolution and 2D max pooling to compress each graph to 2×2 Tianzige-structure with 64 channels. In order to filter noises and blank pixels, we flatten the 2×2 structures and adopt a 1D max pooling to extract glyph vector for each character. The size of glyph vectors is set to 64, which is much smaller than the size of Tianzige-CNN output (1024 dimension).

Different from Glyce that sets image classification task to learn glyph representation, we learn the parameters of CGS-CNN while training whole NER model in domain datasets.

### 3.2 Fusion Stage

We provide a sliding window to slide through both BERT and glyph representations. In sliding window, each slice pair is computed by outer product to capture local interactive features. Then Slice-Attention is adopted to balance the importance of each slice pair and combine them to output fusion representation.

**Out-of-sync Sliding Window.** Sliding window has been applied in multimodal affective computing [14] as mentioned above. The reason for using sliding windows is that directly fusing vectors with outer product would exponentially expand vector size, which increases space and time complexity for subsequent network. However, this method requires the multimodal representations to have the same size, which is not suitable to slide through both BERT vector and glyph vector. Because character representations of BERT have richer semantic information than glyph representations, requiring a bigger vector size. Here we provide an out-of-sync sliding window that can satisfy different vector sizes while keeping the same number of slices.

Assume that we have one Chinese character with character vector defined as $c\_v \in \mathbb{R}^{d^c}$ and glyph vector defined as $g\_v \in \mathbb{R}^{d^g}$. Here $d^c$ and $d^g$ stand for the sizes of two vectors. To keep the same number of the slices of these two vectors after passing through the sliding window, the setting of sliding window needs to meet the following limitation:

$$n = \frac{d^c - k^c}{s^c} + 1 = \frac{d^g - k^g}{s^g} + 1, n \in \mathbb{N}^* \tag{1}$$

Where $n$ is a positive integer, standing for slice number of two vectors; $k^c$ and $s^c$ respectively stand for window size and stride of character vector. $k^g$ and $s^g$ respectively represent window size and stride for glyph vector. The strategy we use to satisfy this condition is to limit the hyper-parameters of sliding window such that $d^c$, $k^c$ and $s^c$ are respectively an integral multiple of $d^g$, $k^g$ and $s^g$.

To get slice pairs, we first calculate the left border index of sliding window at each stride:

$$i \in \{1,2,3\cdots,n\} \tag{2}$$
$$p_{(i)}^c = s^c(i-1) \tag{3}$$
$$p_{(i)}^g = s^g(i-1) \tag{4}$$

Where $p_{(i)}^c$ and $p_{(i)}^g$ represent the boundary index of sliding window respectively for character and glyph vector at the $i$th stride. Then we can obtain each slice during the following formula:

$$c\_s_{(i)} = \left\{c\_v_{(p_{(i)}^c+1)}, c\_v_{(p_{(i)}^c+2)} \cdots, c\_v_{(p_{(i)}^g+k^c)}\right\} \tag{5}$$

$$g\_s_{(i)} = \left\{g\_v_{(p_{(i)}^g+1)}, g\_v_{(p_{(i)}^g+2)} \cdots, g\_v_{(p_{(i)}^g+k^g)}\right\} \tag{6}$$

Where $c\_s_{(i)}$ and $g\_s_{(i)}$ represent the $i$th slices respectively from two vectors; $c\_v_{(p_{(i)}^c+1)}$ stands for the value at $(p_{(i)}^c+1)$th dimension of $c\_v$.

In order to fuse two slices in a local perspective, outer product is adopted to generate an interactive tensor, as shown in the formula:

$$m_i = Outer(c\_s_{(i)}, g\_s_{(i)})$$
$$= \begin{bmatrix} c\_v_{p_{(i)}^c+1} g\_v_{p_{(i)}^g+1}, & \cdots & c\_v_{p_{(i)}^c+1} g\_v_{p_{(i)}^g+k^g} \\ \vdots & \ddots & \vdots \\ c\_v_{p_{(i)}^g+k^c} g\_v_{p_{(i)}^g+1}, & \cdots & c\_v_{p_{(i)}^g+k^c} g\_v_{p_{(i)}^g+k^g} \end{bmatrix} \tag{7}$$

Where $m_i \in \mathbb{R}^{d^c \times d^g}$ stands for fusion tensor of the $i$th slice pair; $c\_v_{p_{(i)}^c+1} g\_v_{p_{(i)}^g+1}$ represent product result between the $p_{(i)}^c+1$th value in $c\_v$ and the $p_{(i)}^g+1$th value in $g\_v$. During outer product, we may obtain all product result among elements from two vectors.

Then we flatten each tensor $m_i$ to vector $m_i' \in \mathbb{R}^{d^c d^g}$. Representation of slices for one character can be represented as:

$$m' = \{m_1', m_2', \ldots m_{n-1}', m_n'\}, m' \in \mathbb{R}^{n \times (k^c k^g)} \tag{8}$$

Where $m'$ contains $n$ fusion vectors of slice pairs. The size of each vector is $k^c k^g$.

**Slice-Attention.** Outer product offers interactive information for character-level representation at the same time generates more noises, as many features are irrelevant. With reference to attention mechanism, we propose the Slice-Attention, which may adaptively quantify the importance of each slice pair and combined them to represent a character. Importance of slice pair can be quantified as:

$$a_i = \frac{exp\left(\sigma(v)\sigma(W^{slice}m'_i + b^{slice})\right)}{\sum_{i=1}^{n} exp\left(\sigma(v)\sigma(W^{slice}m'_i + b^{slice})\right)} \tag{9}$$

Where $a_i$ stands for importance value of the $i$th slice pair; $\sigma$ is Sigmoid function. Sigmoid function here may limit the value range in vectors between 0 and 1, which ensures subsequent dot product computing meaningful. $W^{slice} \in \mathbb{R}^{(k^c k^g) \times (k^c k^g)}$ and $b^{slice} \in \mathbb{R}^{k^c k^g}$ stand for initialized weight and bias. $v \in \mathbb{R}^{(k^c k^g)}$ imitates the query in self-attention [9], which is another initialized weight.

Finally, we fuse the vectors of slice pairs by weighted average computation and obtain fusion vector $f_v$ for a character:

$$f\_v = \sum_{i=1}^{n} a_i m'_i \tag{10}$$

### 3.3 Tagging Stage

We concatenate each vector in character-level before tagging. The final representation of a sentence can be defined as $x = \{x_1, x_2 \dots, x_\tau\}$, where $\tau$ stands for the length of sentence. Then BiLSTM is adopted as sequence encoder and CRF is adopted as decoder for named entity tagging.

**BiLSTM.** LSTM (Long Short Terms Memory) units contain three specially designed gates to control information transmission along a sequence. To encode sequence information of $x$, we use a forward LSTM network to obtain forward hidden state and a backward LSTM network to obtain backward hidden state. Then the two hidden states are combined as:

$$h = \overrightarrow{LSTM}(x) + \overleftarrow{LSTM}(x) \tag{11}$$

Here $h = \{h_1, h_2 \dots, h_\tau\}$ is the hidden representation for characters. We sum the corresponding values between two hidden states to create the $h$.

$$P(y|s) = \frac{exp(\sum_{i=1}^{\tau}(W_{l_i}^{crf} h_i + b_{(l_{i-1}, l_i)}^{crf}))}{\sum_{y'} exp(\sum_{i=1}^{\tau}(W_{l'_i}^{crf} h_i + b_{(l'_{i-1}, l'_i)}^{crf}))} \tag{12}$$

Where $y'$ represents a possible label sequence; $W_{l_i}^{crf}$ represents the weight for $l_i$; and $b_{(l_{i-1}, l_i)}^{crf}$ is the bias from $l_{i-1}$ to $l_i$.

After CRF decoding, we use first-order Viterbi algorithm to find the most probable label sequence for a sentence. Assume that there is a labeled set $\{(s_i, y_i)\}|_{i=1}^{N}$, we minimize the below negative log-likelihood function to train the whole model:

$$L = -\sum_{i=1}^{N} log\left(P(y_i|s_i)\right) \tag{13}$$

## 4 Experiments

In Section 4.1 and Section 4.2, we introduce the situation of datasets we use and some setting of the follow-up experiments. The main experiment result can be found in Section 4.2, where we set a comparison of our model and various SOTA models. FGN we proposed are tested for 10 times in each dataset to compute the average Precision (P), Recall (R), F1-socre (F1). In Section 4.3, we test some main components in FGN and each component is also test for 10 times to compute the average metrics.

### 4.1 Experimental Settings

**Dataset.** Four widely-used NER datasets are chosen for experiments, including OntoNotes 4 [22], MSRA [23], Weibo [24] and Resume [19]. All of these Dataset is annotated with a BMES tagging scheme. Among them, OntoNotes 4 and MSRA are in news domain; Weibo is annotated from Sina Weibo, a social media in China. These three datasets only contain traditional name entities, such as location, personal name and organization. Resume was annotated from personal resumes with 8 types of named entities.

**Hyper-Parameter Setting.** We use dropout mechanism for both character representation and glyph representation. Dropout rate of CGS-CNN is set to 0.2 and the one of radical self-attention is set to 0.5. The hidden size of LSTM is set to 764 and the dropout rate of LSTM is set to 0.5. We used the Chinese BERT which was pre-trained by Google[2]. Following the default configuration, output vector size of each character is set to 764. Character graphs we used are collected from *Xinhua Dictionaries*[3] with the number of 8630. We covert these graphs to 50×50 gray-scale graph. As mentioned in Section 3.2, window size and stride in sliding window of character vector are respectively an integer multiple of the ones for glyph vectors. Thus, we set size and stride of the former to 96 and 8, and the later to 12 and 1 according to empirical study. Adam is adopted as optimizer for both BERT fine-tuning and NER model training. Learning rates for fine-tuning condition and training condition are different. The former one is set to 0.00001, and the latter one is set to 0.002.

### 4.2 Main Result

Table 1 and Table 2 show some detailed statistics of FGN, which is compared with other SOTA models on four NER datasets. Here FGN represents the proposed glyph model with LSTM-CRF as tagger; Lattice LSTM [19] and WC-LSTM [20] are the SOTA model without BERT, combining both word embedding and character embedding. BERT-LMCRF represent the BERT model with BiLSTM-CRF as NER tagger. Glyce [2] is the SOTA BERT-based glyph network as mentioned earlier. GlyNN [11] is another SOTA BERT-based glyph network. Especially, we select the average F1 of GlyNN for comparison as we also adopt the average F1 as metric. For other baselines, we select their result shown in trial, as they have not illustrated whether they used the average F1 or not.

---

[2] https://github.com/google-research/bert
[3] http://zidian.aies.cn/

As can be seen, FGN outperforms other SOTA models in all four datasets. Compared with BERT-LMCRF, F1 of FGN obtains obvious boosts of 3.13%, 2.88%, 1.01% and 0.84% respectively on Weibo, OntoNote 4, MSRA and Resume. Further, FGN outperformed some SOTA glyph-based NER model like Glyce and GlyNN. However, FGN did not achieve significant improvement on Resume and MSRA dataset as BERT-LMCRF can already recognize most of the entities on these two datasets. In fact, the datasets Weibo and OntoNote4 are more difficult for NER, as the entity types and entity mentions are more diverse. For example, some interesting and extraordinary entity words in Weibo and OntoNote4 like "铼德" (company name) and "啊滋猫" (milk tea shop), which were successfully identified only by FGN. We guess the reason is because the character "铼" contain the radical "钅" which means "metal" and the character "滋" contains the radical "氵" which means "water". These radicals are related to the products of their companies. In fact, this phenomenon is common in various Chinese entities including company, personal name and location, which are deeply influenced by the naming culture of Chinese people. Combined the contextual information with the above glyph information, FGN may capture extra feature to recognize some extraordinary named entities in some cases.

Table 1. Detailed statistics of FGN on Weibo and OntoNote 4

| Model | Weibo | | | OntoNote 4 | | |
|---|---|---|---|---|---|---|
| | P | R | F1 | P | R | F1 |
| Lattice-LSTM | 53.04 | 62.25 | 58.79 | 76.35 | 71.56 | 73.88 |
| WC-LSTM | 52.55 | 67.41 | 59.84 | 76.09 | 72.85 | 74.43 |
| BERT-LMCRF | 66.88 | 67.33 | 67.12 | 78.01 | 80.35 | 79.16 |
| Glyce | 67.68 | 67.71 | 67.70 | 80.87 | 80.40 | 80.62 |
| GlyNN | N/A | N/A | 69.20 | N/A | N/A | N/A |
| FGN | **69.02** | **73.65** | **71.25** | **82.61** | **81.48** | **82.04** |

Table 2. Detailed statistics of FGN on Resume and MSRA.

| Model | Resume | | | MSRA | | |
|---|---|---|---|---|---|---|
| | P | R | F1 | P | P | F1 |
| Lattice-LSTM | 93.57 | 92.79 | 93.18 | 93.57 | 92.79 | 93.18 |
| WC-LSTM | 95.27 | 95.15 | 95.21 | 94.58 | 92.91 | 93.74 |
| BERT-LMCRF | 96.12 | 95.45 | 95.78 | 94.97 | 94.62 | 94.80 |
| Glyce | **96.62** | 96.48 | 96.54 | **95.57** | 95.51 | 95.07 |
| GlyNN | N/A | N/A | 95.66 | N/A | N/A | 95.21 |
| FGN | 96.49 | **97.08** | **96.79** | 95.45 | **95.81** | **95.64** |

### 4.3 Ablation Study

Here we discuss the influences of various settings and components in FGN. The comp-onents we investigate contain: CNN structure, named entity tagger and fusion method. Weibo dataset is used for these illustrations.

**Effect of CNN structure.** As shown in Table 3, we investigate the performances of various CNN structures while keeping other settings of FGN unchanged. In this table, "*2d*" represents the CGS-CNN with no 3D convolution layer. "*avg*" represents that 1D max pooling in CGS-CNN is replaced by 1D average pooling. 2D CNN represents the CNN structure with only 2D convolution and 2D pooling layers. Tianzige-CNN is proposed from Glyce.

As can be seen, the common 2D-CNN structure obtains the worse result, as it completely overlooks the information of Tianzige structure and neighbor character glyph. Comparing with Tianzige-CNN, using CGS-CNN introduces a boost of 0.66% in F1,as CGS-CNN may capture interactive information between the character glyph. Compared with 2D convolution, Using FGN with 3D convolution introduces a boost of 1.14% in F1, which confirmed the benefit from adjacent glyph information of phrases or words. Otherwise, max pooling works better than average pooling when capture feature in Tianzige structure. As mentioned earlier, max pooling here may filter some blank pixels and noises in character graphs.

**Table 3.** Performances of various CNN structures on Weibo dataset.

| CNN-type | P | R | F1 |
|---|---|---|---|
| CGS-CNN$^{2d}$ | 68.56 | 71.45 | 70.01 |
| CGS-CNN$^{avg}$ | 69.13 | 71.35 | 70.22 |
| 2D-CNN | 67.75 | 72.45 | 69.93 |
| Tianzige-CNN | **70.94** | 70.24 | 70.59 |
| CGS-CNN | 69.02 | **73.65** | **71.25** |

**Effect of Named Entity Tagger.** Some widely-used sequence taggers are chosen to replace BiLSTM-CRF in FGN for discussion. Table 4 shows the performances of various chosen taggers. As can be seen, methods that based on LSTM and CRF outperform Transformer [9] encoder in NER task. In fact, Most of the SOTA NER methods [11, 19, 20] prefer to use BiLSTM rather than Transformer as their sequence encoder. Compared with only CRF, LSTM-CRF introduces a boost of 0.43% in F1. In addition, bidirectional LSTM introduces a further boost of 0.56% in F1. In this experiment, LSTM-CRF performed better than Transformer in NER task.

**Table 4.** Performances of various taggers on Weibo dataset.

| tagger-type | P | R | F1 |
|---|---|---|---|
| CRF | 7044 | 70.10 | 70.26 |
| LSTM-CRF | 70.77 | 70.60 | 70.69 |
| BiLSTM-CRF | 69.02 | **73.65** | **71.25** |
| Transformer | **72.14** | 66.08 | 68.98 |

**Effect of Fusion Method.** We investigate the performances of different setting in fusion stage as shown in Table 5. In this table, "concat" represents concatenating glyph and BERT representation without any fusion; "no freeze" represents FGN with trainable BERT; "avg pool" and "max pool" represent that Slice-Attention in FGN is respectively replaced by pooling or max pooling. In addition, we reset the window size to (196, 16), (48, 4) and the stride to (24, 2) in sliding window respectively for character and glyph representations to test the FGN.

Compared to directly concatenating vectors from glyph and BERT, FGN introduces a boost of 0.82% in F1, which confirms the effectiveness of our fusion strategy. FGN with the strategy of fine-tuning and freezing BERT in different stages outperforms the FGN with a trainable BERT. We consider is because that fine-tuning BERT only requires minimal gradient values when updating the BERT parameters, but LSTM-CRF need to set a larger learning rate to adjusting the initialized parameter with suitable gradient values. Using Slice-Attention outperforms using average pooling or max pooling in FGN, as Slice-Attention adaptively balances information of each slices and pooling layer only filter information statically. Otherwise, sliding window with the setting in Section 4.1 slightly outperforms the ones with other hyper-parameter settings.

Table 5. Performances of different fusion settings on Weibo dataset.

| fusion-type | P | R | F1 |
|---|---|---|---|
| concat | 69.13 | 71.35 | 70.43 |
| no freeze | 66.92 | **74.87** | 70.67 |
| avg pool | 69.00 | 73.61 | 70.11 |
| max pool | 69.60 | 71.40 | 70.64 |
| w(196, 16) | **70.58** | 71.10 | 70.84 |
| w(48, 4) | 70.25 | 71.22 | 70.73 |
| s(24, 2) | 69.07 | 73.00 | 70.98 |
| FGN | 69.02 | **73.65** | **71.25** |

## 5 Conclusion

In this paper, we proposed the FGN for Chinese NER. In FGN, a novel CNN structure called CGS-CNN was applied to capture both glyph information and interactive information between the neighboring graphs. Then a fusion method with out-of-sync sliding window and Slice-Attention were adopted to fuse the output representations from BERT and CGS-CNN, which may offer extra interactive information for NER tasks. Experiments are conducted on four NER datasets, showing that FGN with LSTM-CRF as tagger obtained SOTA performance on four datasets. Further, influences of various settings and components in FGN are discussed during ablation study.

## Acknowledgments

This work was supported by the National Natural Science Foundation of China(No. 61572145) and the Major Projects of Guangdong Education Department for Foundation Research and Applied Research (No. 2017KZDXM031). The authors would like to thank the anonymous reviewers for their valuable comments and suggestions.